# Neural Network based Deep Transfer Learning for Cross-domain Dependency Parsing

Zhentao Xia[1], Likai Wang[1], Weiguang Qu[1], Junsheng Zhou[1], Yanhui Gu[1]

[1]Department of Computer Science and Technology, Nanjing Normal University, Nanjing, China
{zt950607,wlk_1181772577}@foxmail.com, wggu_nj@163.com
{zhoujs,gu}@njnu.edu.cn

**Abstract.** In this paper, we describe the details of the neural dependency parser submitted by our team to the NLPCC 2019 Shared Task of Semi-supervised domain adaptation subtask on Cross-domain Dependency Parsing. Our system is based on the stack-pointer networks(STACKPTR). Considering the importance of context, we utilize self-attention mechanism for the representation vectors to capture the meaning of words. In addition, to adapt three different domains, we utilize neural network based deep transfer learning which transfers the pre-trained partial network in the source domain to be a part of deep neural network in the three target domains (product comments, product blogs and web fiction) respectively. Results on the three target domains demonstrate that our model performs competitively.

## 1     Introduction

The goal of the NLPCC 2019 Shared Tasks on Cross-domain Dependency Parsing is to predict the optimal dependency tree that can adapt three different domains from source domain.

Dependency parsing is an important component in various natural language processing systems for semantic role labeling[4], relation extraction [5], and machine translation [6]. There are two dominant approaches to dependency parsing: graph-based algorithms [9] and transition-based algorithms [10].With the surge of web data, cross-domain parsing has become the major challenge for applying syntactic analysis in realistic NLP systems.

In this paper, we describe the details of our dependency parser system submitted to the NLPCC 2019 Shared Task of Semi-supervised domain adaptation subtask on Cross-domain Dependency Parsing. Our system is based on the stack-pointer network dependency parser [7]. The model has a pointer network as its backbone, and is equipped with an internal stack to maintain the order of head words in tree structures. To capture the context of sentences, we obtain word representations by self-attention mechanism [8].

Existing studies on dependency parsing mainly focus on the in-domain setting, where both training and testing data are drawn from the same domain. How to build dependency parser that can learn across domains remains an under-addressed problem. In our work, we study cross-domain dependency parsing. We model it as a domain adaptation problem, where we are given one source domain and three target domains, and the core task is to adapt a dependency parser trained on the source domain to the target domain. Inspired by the recent success of transfer learning in many natural language processing problems, we utilize neural network based deep transfer learning for cross-domain dependency parsing. It refers to the reuse the partial network that pretrained in the source domain, including its network structures and connection parameters, transfer it to be a part of deep neural network which used in target domain [12].

The rest of this paper is organized as follows. Section 2 gives a description of our parser system, including the system framework and stack-pointer network with self-attention mechanism for dependency parsing. In Section 3, we describe neural network based deep transfer learning for domain adaptation. In Section 4, we list our experiments and discuss results.

## 2   System Overview

The model architecture of our dependency parsing system, which uses STACKPTR parser [7] as its backbone. The structure of the system is shown in Fig. 1.

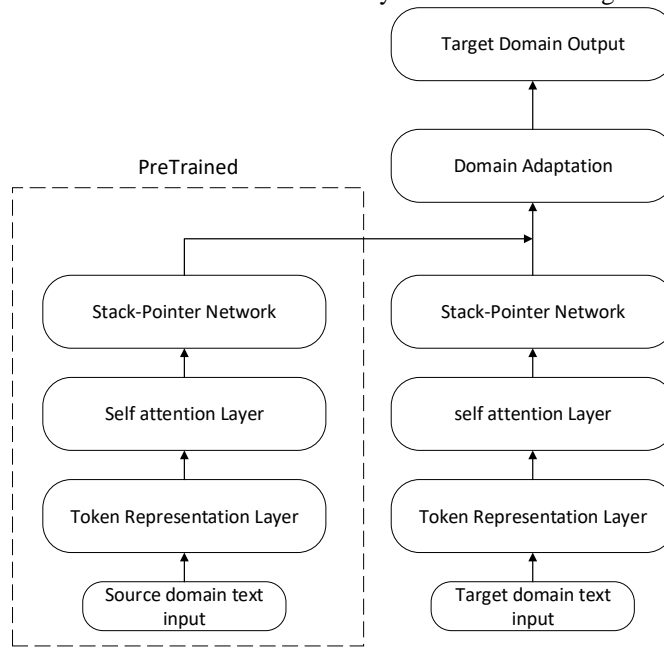

**Fig. 1.** The structure of the system

The system mainly contains four components: the token representation layer, the self attention layer, the stack-pointer network architecture for dependency parsing and the domain adaptation with deep transfer learning. We describe the four sub-modules in the following sections in details.

### 2.1 Token representation

Let an input sentence is denoted as $S = \{w_1, w_2, \ldots, w_n\}$, where n is the number of words. The token representation has three parts:
**Word-level Embedding.** We transform each word into vector representation by looking up pre-trained word embedding matrix $W_{word} \in R^{d_w \times |V|}$, where $d_w$ is the dimension of the vectors and |V| is the size of vocabulary.
**Character-level Embedding.** To encode character-level information of a word into its character-level representation, we run a convolution neural network on the character sequence of $w_i$. Then the character-level embedding vector is concatenated with the word-level embedding vector for each word representation.
**POS Embedding.** To enrich word representation information, we also use POS embedding. Finally, the POS embedding vectors are concatenated with word embedding vectors as context information inputs $X = \{x_1, x_2, \ldots, x_n\}$ to feed into next layer.

### 2.2 Stack-pointer networks

Ma et al. [7] implement a new neural network architecture called stack-pointer networks (STACKPTR) for dependency parsing. STACKPTR parser has a pointer network as its backbone. This model is equipped with an internal stack to maintain the order of head words in tree structures.

The model firstly reads the whole sentence and encodes each word with BiLSTMs into the encoder hidden state $e_i$.

The decoder implements a top-down, depth-first transition system. At each time step t, the decoder receives the encoder hidden state $e_i$ of the word $w_i$ on top of the stack to generate a decoder hidden state $d_t$ and computes the attention vector $a^t$ using the following equation:

$$v_i^t = \text{score}(d_t, s_i) \quad (1)$$
$$a^t = \text{softmax}(v^t) \quad (2)$$

For attention score function, the model adopt the biaffine attention mechanism described in Dozat et al. [11]. The pointer network returns a position p according to the highest attention score in $a^t$ and generate a new dependency arc $w_i \rightarrow w_p$ where $w_p$ is considered as a child of $w_i$. Then the parser pushes $w_p$ onto the stack. If the parser pointers $w_i$ to itself, then $w_i$ is considered to have found all its children. Finally the parser goes to the next step and pops $w_i$ out of stack. The parsing process ends when only the root contains in the stack.

A dependency tree can be represented as a sequence of top-down paths $p_1, \ldots, p_k$, where each path $p_i$ corresponds to a sequence of words $\$, w_{i,1}, w_{i,2}, \ldots, w_{i,li}$ from the root to a leaf. The STACKPTR parser is trained to optimize the probability:

$$P_\theta(y|x) = \prod_{i=1}^{k} P_\theta(p_i|p_{<i}, x) = \prod_{i=1}^{k} \prod_{j=1}^{l_i} P_\theta(w_{i,j}|w_{i,<j}, p_{<i}, x)$$

Where θ represents model parameters, $p_{<i}$ stands for previous paths already explored, $w_{i,j}$ denotes the jth word in path $p_i$ and $w_{i,<j}$ represents all the previous words on $p_i$.

### 2.3 Self Attention Layer

In order for the representation vectors to capture the meanings of words considering the context, we employ the self-attention, a special case of attention mechanism [8]. We adopt the multi-head attention formulation, one of the methods for implementing self-attention. Fig.2 illustrates the multi-head attention mechanism.

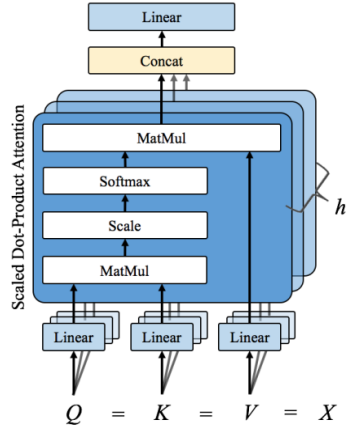

**Fig. 2.** The multi-head attention mechanism

Given a matrix of n vectors, query Q, key K and value V, The formulation of multi-head attention is defined by the follows:

$$\text{Attention}(Q, K, V) = \text{softmax}\left(\frac{QK^T}{\sqrt{d_w}}\right) V \qquad (3)$$

$$\text{MultiHead}(Q, K, V) = W^M[head_1; \ldots, head_r] \qquad (4)$$

$$head_i = \text{Attention}(W_i^Q Q, W_i^K K, W_i^V V) \qquad (5)$$

Where [;] indicates row concatenation and r is the number of heads. The weights $W^M \in R^{d_w \times d_w}, W_i^Q \in R^{\frac{d_w}{r} \times d_w}, W_i^K \in R^{\frac{d_w}{r} \times d_w}, W_i^V \in R^{\frac{d_w}{r} \times d_w}$ are learnable parameters for linear transformation. As a result, the output of self attention layer is the sequence of representations whose include informative factors in the input sentence as model input.

## 3   Domain Adaptation

From our study, insufficient target domain training data is an inescapable problem. Transfer learning relaxes the hypothesis that the training data must be independent, which motivates us to use transfer learning in our work. The model in target domain is not need to trained from scratch, which can reduce the demand of training data in the target domain and can learn enough knowledge from source domain.

In our work, the stack-pointer network is trained in source domain with training dataset. Second, we reuse the partial network that pre-trained in the source domain, including its network structure and connection parameters, transfer it to be a part of deep neural network which used in target domain. Finally, the transferred sub-network may be updated in fine-tune strategy. The overview of neural network based transfer learning is shown in Fig.3.

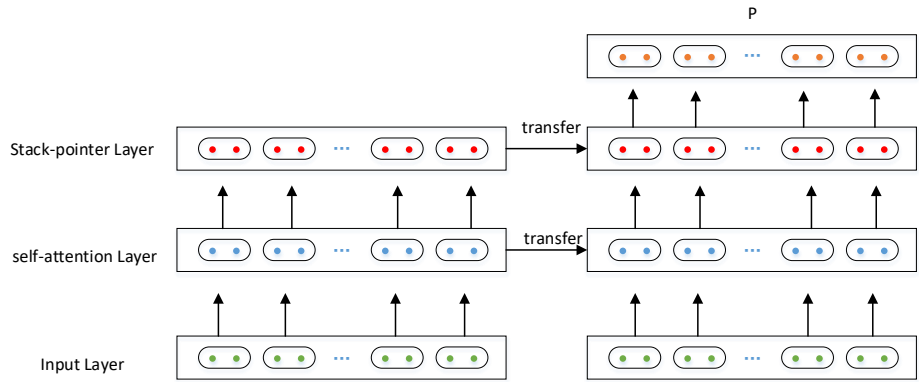

**Fig. 3.** The neural network based transfer learning

Concretely, we retain the parameters of encoder and decoder and abandon the parameters of the biaffine attention mechanism which trained in source domain. We retrain a new biaffine attention score and fine-tune the parameters of the whole network for each of the three target domains.

Besides, to avoid insufficient target domain dataset and can learn enough knowledge from source domain, we retain the self-attention parameters which trained in source domain and fine-tune for target domains since corpus of source domain is larger than three target domains. From our cognition, although target datasets are different areas, the understanding of semantic in Chinese is the same. Transfer learning can also help improve the network efficiently with limited amount of training data in target domains.

## 4 Experiments

### 4.1 Data and Evaluation Metrics

We use the data provided by official. The official provides one source-domain and three target-domain datasets. Source domain data are selected from HLT-CDT and PennCTB treebanks (BC). For Target domains, data are selected from Taobao as comments products (PC), Taobao headlines as product blogs (PB) and web fiction "Zhuxian" (ZX) respectively. Table 1 shows data distribution for the task.

The official references the HLT-CDT and UD annotation guidelines, and developed a detailed annotation guideline that aims to fully capture Chinese syntax and tries to guarantee inter-annotator consistency and facilitate model learning. The guideline includes 20 dependency labels. In order to reduce annotation cost, they adopt the active learning procedure based on partial annotation [2]. All training datasets are automatically complemented into high-quality full trees [3].

**Table 1. Source domain and Target domain data distribution**

|  |  | Train | Dev | Test |
|---|---|---|---|---|
| Source Domain | Balanced Corpus (BC) | 16.3k | 1k | 2k |
| Target Domains | Comments Products (PC) | 6.2k | 1.3k | 2.6k |
|  | Product Blogs (PB) | 5.1k | 1.3k | 2.6k |
|  | The web fiction (ZX) | 1.6k | 0.5k | 1.1k |

We adopt the official evaluation metric, which is based on the standard labeled attachment score (LAS, percent of words that receive correct heads and labels). The official average the three target domain LAS directly to determine the final ranking.

### 4.2 Settings

We use the pre-trained weights of the publicly available Glove model [13] to initialize word embedding in our model, and use random initialization for POS embedding.

We employ Adam method to optimize our model. Following Ma et al. [7], we apply dropout training to mitigate overfitting. The hyper-parameters in our model are shown in following Table 2.

**Table 2. Hyper-parameters setting**

| Hyper-parameters | Description | Values |
|---|---|---|
| $d_w$ | Size of Word Embedding | 300 |
| char_dim | Dimension of character embedding | 50 |
| pos_dim | Dimension of POS embedding | 50 |
| r | Number of heads | 4 |
| $d_h$ | Number of hidden units in RNN | 256 |
| batch_size | Number of sentences in each bacth | 64 |
| num_filters | Number of filters in CNN | 50 |
| learning_rate | Learning rate | 0.001 |
| decay_rate | Decay rate of learning rate | 0.75 |
| p_rnn | Dropout rate for RNN | 0.5 |
| p_in | Dropout rate for input embedding | 0.5 |
| p_out | Dropout rate for output layer | 0.5 |

### 4.3 Result

In order to analyze our model, we compare with baseline model STACKPTR [7]. We conduct an ablation experiment on our model to examine the effectiveness of self-attention and transfer learning components. The overall results on development datasets are illustrated in Table 3. The official result [14] on test datasets is provided under the NLPCC 2019 shared task on Cross-domain Dependency Parsing website.

Table 3. The overall results on development datasets

| Model | PC LAS | PB LAS | ZX LAS | Average LAS |
|---|---|---|---|---|
| STACKPTR | 61.1 | 74.8 | 74.6 | 70.2 |
| STACKPTR+multi-head attention | 60.9 | 75.5 | 75.1 | 70.5 |
| STACKPTR+transfer learning | 61.9 | 75.7 | 75.4 | 71 |
| STACKPTR+multi-headattention+transfer learning | 62.6 | 76.9 | 76.3 | 71.9 |

The results indicate that our model with self-attention mechanism and transfer learning can promote the performance in three different target domains. Recalling the model architecture, the multi-head attention can capture different features for cross-domain dependency parsing and transfer learning can also help improve the network efficiently with limited amount of training data in target domains.

## 5 Conclusions

In this paper, we present the neural dependency parser submitted by our team to the NLPCC 2019 Shared Task of Semi-supervised domain adaptation subtask on Cross-domain Dependency Parsing, which includes self-attention mechanism and neural network based deep transfer learning. The results suggested that using self-attention and transfer learning is a way to achieve competitive cross-domain parsing performance. Our model on comments product domain does not perform well. We will continue to improve our system in future work.